\documentclass{article}

\PassOptionsToPackage{numbers}{natbib}

\usepackage[preprint]{neurips_2024}

\usepackage[utf8]{inputenc} 
\usepackage[T1]{fontenc}    
\usepackage[pagebackref,breaklinks,colorlinks]{hyperref}
\usepackage{url}            
\usepackage{amsfonts}       
\usepackage{nicefrac}       
\usepackage{microtype}      
\usepackage{xcolor}         
\usepackage{float}          
\usepackage{graphicx}
\usepackage[absolute,overlay]{textpos}
\usepackage{tikz}

\usepackage{amsmath}
\usepackage{amsfonts}
\usepackage{amssymb}
\usepackage{wrapfig}
\usepackage{subcaption}
\usepackage{multirow}
 \usepackage{mathtools} 

\usepackage{verbatim}

\usepackage{anyfontsize}

\usepackage{microtype}
\usepackage{graphicx}
\usepackage{booktabs} 
\usepackage{multirow}
\usepackage{amsmath,amssymb}
\usepackage{booktabs}
\usepackage{caption,subcaption}

\usepackage{xcolor}
\definecolor{mygreen}{HTML}{3cb44b}
\definecolor{skyblue}{HTML}{beffff}
\definecolor{lightgreen}{HTML}{90ee90}

\usepackage{color, colortbl}

\definecolor{emerald}{rgb}{0.31, 0.78, 0.37}

\usepackage{tcolorbox}
\usepackage{enumitem}
\setitemize{itemsep=10pt,topsep=0pt,parsep=0pt,partopsep=0pt}
\usepackage{colortbl}

\usepackage{xcolor}
\definecolor{mygreen}{HTML}{3cb44b}
\colorlet{myyellow}{green!10!orange!90!}
\makeatletter

\usepackage{tikz}
\usetikzlibrary{arrows,shapes,snakes,automata,backgrounds,fit,petri}
\usepackage{adjustbox}

\newcommand{\RN}[1]{%
	\textup{\lowercase\expandafter{\it \romannumeral#1}}%
}
\usepackage{tabu}

\newcommand{\beq}{\vspace{0mm}\begin{equation}}
\newcommand{\eeq}{\vspace{0mm}\end{equation}}
\newcommand{\beqs}{\vspace{0mm}\begin{eqnarray}}
\newcommand{\eeqs}{\vspace{0mm}\end{eqnarray}}
\newcommand{\barr}{\begin{array}}
\newcommand{\earr}{\end{array}}

\newcommand{\Xmat}[0]{{{\bf X}}}

\newcommand{\xv}{\boldsymbol{x}}

\newcommand{\thetav}{\boldsymbol{\theta}}

\usepackage{color, colortbl}
\definecolor{Gray}{gray}{0.93}

\usepackage{lipsum}

\usepackage{pifont}

\usepackage{makecell}

\usepackage{xcolor,amsmath}
\usepackage[linesnumbered,ruled,vlined]{algorithm2e}
\DontPrintSemicolon

\usepackage{xcolor}
\definecolor{mygreen}{HTML}{3cb44b}

\SetKwComment{Comment}{\color{green!50!black}\# }{}

\SetKwProg{Function}{def}{:}{}

\SetKwProg{For}{for}{:}{}
\SetKwProg{If}{if}{:}{}
\newcommand{\VarSty}[1]{\textnormal{\ttfamily\color{blue!90!black}#1}\unskip}

\usepackage{xcolor}

\newcommand{\shortname}{SQ}

\newcommand{\longshortname}{{\bf{Socratic Questioning (SQ)}}}

\title{Socratic Questioning: Learn to Self-guide Multimodal Reasoning in the Wild}
\title{\begin{tikzpicture}[remember picture,baseline=(current bounding box.base)]
\node[anchor=base, xshift=0.2cm, yshift=-0.15cm]{\includegraphics[width=1cm]{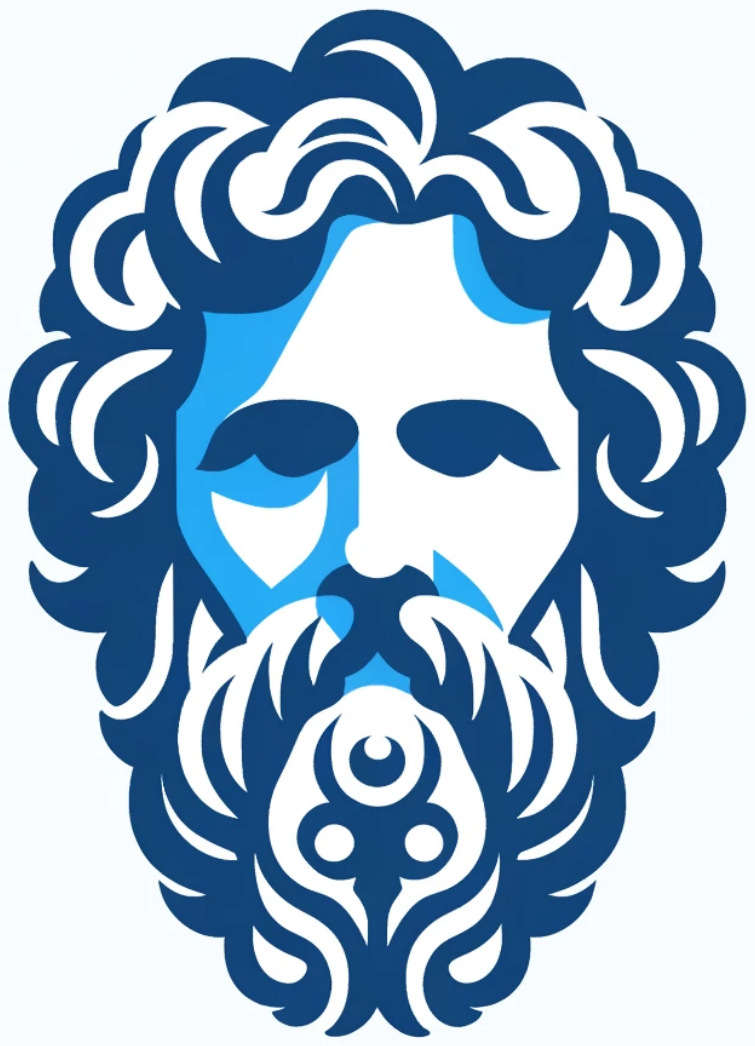}};
\end{tikzpicture}Socratic Questioning: Learn to Self-guide Multimodal Reasoning in the Wild}

\author{
  Wanpeng Hu\textsuperscript{1}\thanks{Equal contribution. Names are sorted randomly.}, 
  Haodi Liu\textsuperscript{2}\footnotemark[1], 
  Lin Chen\textsuperscript{1}, 
  Feng Zhou\textsuperscript{1}, \\
  \textbf{Changming Xiao}\textsuperscript{2}, 
  \textbf{Qi Yang}\textsuperscript{2}, 
  \textbf{Changshui Zhang}\textsuperscript{2}\thanks{Corresponding author: zcs@mail.tsinghua.edu.cn.} \\
  \textsuperscript{1}Aibee Inc ~~~~~~ \textsuperscript{2}Tsinghua University \\
  \href{https://github.com/aibee00/SocraticQuestioning}{https://github.com/aibee00/SocraticQuestioning}
}

\begin{document}

\maketitle

\begin{abstract}
Complex visual reasoning remains a key challenge today. Typically, the challenge is tackled using methodologies such as Chain of Thought (COT) and visual instruction tuning. However, how to organically combine these two methodologies for greater success remains unexplored. Also, issues like hallucinations and high training cost still need to be addressed. In this work, we devise an innovative multi-round training and reasoning framework suitable for lightweight Multimodal Large Language Models (MLLMs). Our self-questioning approach heuristically guides MLLMs to focus on visual clues relevant to the target problem, reducing hallucinations and enhancing the model's ability to describe fine-grained image details. This ultimately enables the model to perform well in complex visual reasoning and question-answering tasks. We have named this framework \longshortname{}. To facilitate future research, we create a multimodal mini-dataset named {\bf{CapQA}}, which includes 1k images of fine-grained activities, for visual instruction tuning and evaluation, our proposed SQ method leads to a 31.2\% improvement in the hallucination score. Our extensive experiments on various benchmarks demonstrate \shortname's remarkable capabilities in heuristic self-questioning, zero-shot visual reasoning and hallucination mitigation. Our model and code will be publicly available.
\end{abstract}

\section{Introduction}

Effective visual reasoning and question answering in complex scenarios are highly valuable, as they provide accurate and in-depth insights that can be crucial in practical applications. Currently, visual reasoning and question answering in complex scenes remain a significant challenge. Researchers are actively developing models, making training and fine-tuning datasets, and creating evaluation benchmarks to improve performance in this area.\\

Chain of Thought (COT) and visual instruction tuning are the common methods used to tackle complicated visual reasoning and question answering tasks. Both methods have developed over time to become effective and mature, but how to organically combine them for complementary advantages remains an area worth exploring. At the same time, both methods face challenges such as hallucinations and high training costs.\\
\paragraph{\longshortname{}:}In this paper, we propose an innovative multi-round training and reasoning framework compatible with lightweight Multimodal Large Language Models (MLLMs). Our method is named \longshortname{}: Facing a main problem, \shortname{} uses heuristic, continuous, and in-depth self-questioning to encourage deeper and more comprehensive thinking. This process helps identify errors, broaden perspectives, spark inspiration, and ultimately lead to discovering the truth. \shortname{} elegantly integrates the ideas and techniques of \textbf{Chain of Thought (CoT)} and \textbf{Visual Instruction Tuning}, combining the advantages of both while effectively reducing hallucinations and saving annotation and training costs. An illustration of how \shortname{} works is shown at Figure~\ref{fig:sq_example}.\\
\paragraph{Chain-of-Thought\&Visual Instruction Tuning:}Extensive research has shown that simulating the step-by-step reasoning process of humans can significantly enhance the performance of LLMs on reasoning tasks. Consequently, the Chain of Thought (CoT) approach was proposed and has become a standard method for addressing complex reasoning problems, later extended to multimodal domain. Meanwhile, great works like LLAVA\cite{LLAVA}, LLAVA-1.5\cite{LLAVA_L}, InstructBLIP\cite{InstructBLIP}, Qwen-vl\cite{Qwen-vl} have demonstrated the great success of visual instruction tuning in creating a general-purpose model that can effectively follow multimodal instructions, align with human intents and preferences, and accomplish zero-shot generalizations on unseen data. \\

Socratic Questioning(SQ), a heuristic self-guiding approach, represents a significant refinement and innovation of the current CoT methodology. SQ tackles a complicated visual reasoning and question answering problem in four steps: \\

1. {\it Self-ask}: Figure out what fine-grained information are needed for our reasoning tasks by coming up with some questions to ask itself. \\
2. {\it Self-answer}: Acquire the demanded fine-grained information visually grounded in the image by answering the previously self-asked questions. \\
3. {\it Consolidate \& Organize}: Produce the detailed description of image by coherently consolidating and organizing the information contained in the generated Q\&A pairs.\\
4. {\it Summarize \& Condense}: Produce the summarized caption retaining the core elements by summarizing the information most relevant to our reasoning tasks and condensing the detailed description. \\

We organize the prompts(with image), self-asked questions, corresponding answers,  detailed descriptions, and summarized captions into an instructional conversation format, thereby creating a multimodal mini-data named \textbf{CapQA} for visual instruction tuning the MLLMs. Despite being fine-tuned only on the tiny CapQA dataset, the MLLM given by SQ method has shown impressive zero-shot performance on multiple visual reasoning and question-answering benchmarks that test comprehensive knowledge and recognition abilities, demonstrating its versatility as well as the success of our refinement and innovation of the CoT methodology.\\
\paragraph{Hallucination\&Training Cost:}The issue of hallucinations has consistently accompanied the development of Large Language Models (LLMs), posing a significant challenge to their reliability. Remarkably, our experiments demonstrate that \longshortname{} effectively reduces hallucinations without incurring additional cost like complicated architectures, larger modules and extra data processing. Additionally, \longshortname{} can be widely adapted to various MLLMs, particularly lightweight ones, thus enabling us to avoid substantial training cost. As is well know, CoT methods and visual instruction tuning methods require annotations of rationales and instructional conversations respectively. In order to save the huge cost of manual annotating, researchers have leveraged LLMs to automatically generate annotations of data. Following this good practice, we utilize GPT-4v \cite{GPT-4V} to generate our annotations. \\

In this paper, we present \longshortname{}, an flexible, reliable and effective framework for visual reasoning and question answering in complex scenes. It draws on the principles of the Socratic Questioning, guiding oneself through heuristic questioning to better understand the problem and its context, ultimately providing an informative and insightful description and caption with particularly few hallucinations. Our paper makes the following key contributions:\\

1. We propose an innovative visual reasoning framework \longshortname{} that cleverly integrate the advantages of CoT and visual instruction tuning while effectively reducing hallucinations and training costs.\\

2. We create a mini-dataset \textbf{CapQA} for fine-tuning and evaluations. Despite its small size, \textbf{CapQA} successfully endows MLLMs with the capabilities of heuristic self-questioning, reliable key information retrieval, and zero-shot visual reasoning. It also serves as a good benchmark for visual reasoning and question answering on fine-grained human activity.\\

3. We evaluate our framework on various benchmarks for visual reasoning and hallucinations. Additionally, GPT4\cite{GPT4} is leveraged to help assess the quality of self-asked questions and hallucination levels of descriptions. The extensive experiments strongly support our claims about \longshortname{}.

\begin{figure}[!t]
\centering  
\vspace{-4mm}
\includegraphics[height=11cm]{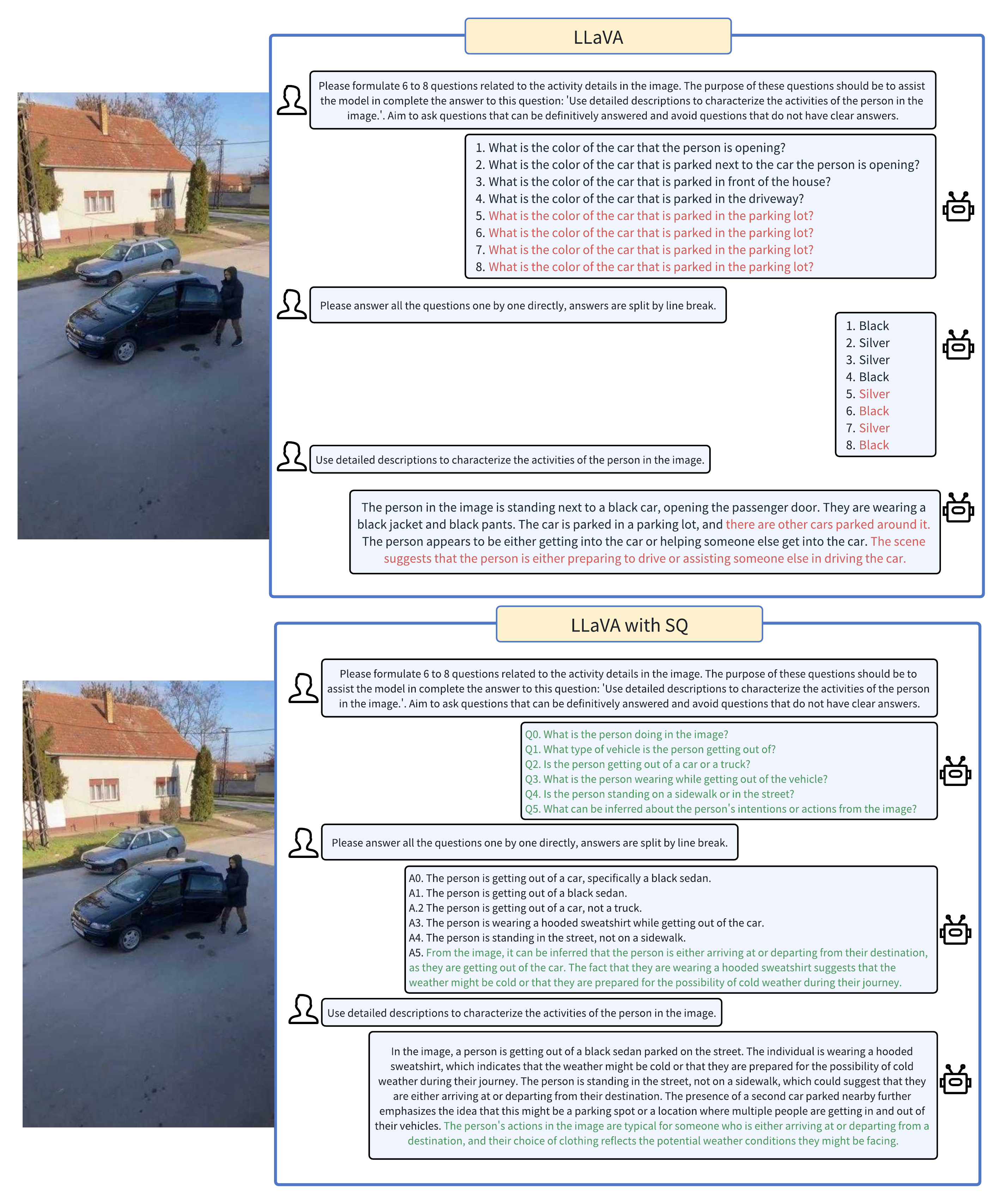}
\vspace{-1mm}
\caption{Comparison of Questions Generation on LLaVA-with-SQ and LLaVA.}
\label{fig:comparision_questions_gen}  
    \vspace{-3mm}
\end{figure}

\vspace{-2mm}
\section{Related Work}
\vspace{-2mm}

\subsection{Multimodal Chain of Thought}
 MM-CoT\cite{AlexSmola} first proposed a two-stage reasoning framework where an LLM initially processes image-text data to obtain a rationale, and then the rationale is fed into the LLM to obtain the final answer. Some subsequent works concentrate on the better alignment and fusion of language and vision modalities. DPMM-CoT\cite{Liqi} leverages the idea and architecture of T2I stable diffusion model to flexibly adjust visual feature extraction according to problem prompts. Additionally, some research focuses on using graph data to encode people, objects, and their mutual relationships. This approach aims to capture more fine-grained information from images, thereby enhancing the benefits of multimodal CoT from the visual modality and reducing hallucinations. KAM-CoT\cite{Mondal} harnesses graph neural networks to process and encode the Knowledge Graph produced from each image, while CCoT\cite{CCoT} delicately prompts LLM to generate a scene graph in Json dictionary format from each image. \\

Substantial efforts have also been made in reducing the annotation costs incurred by the huge demand of training data. Utilizing AI system to automatically generate data is a common and natural practice. T-SciQ\cite{TSciQ} automatically generates teaching data containing question-answer-COT (Chain of Thought) from LLMs and the teaching data will be applied for future finetuning. CURE\cite{CURE}devises an LLM-Human-in-the-Loop pipeline to semi-automatically generate training data and explicitly models fine-grained reasoning chains composed of coherent sub-questions and corresponding answers, from which we can abstract higher-level rationales. Moreover, there are works devoted to uncover and unleash LLMs' judgement and self-guiding capabilities to tackle problems more effectively. DDCoT\cite{GeZheng} guides the LLMs to decompose the main problem and carefully distinguish which parts can be answered using its own knowledge and which parts require information provided by a visual recognition model. Ultimately, the LLM and the visual recognition model each performs their respective tasks, working together to form a complete problem-solving reasoning chain.\\
\subsection{Hallucination mitigation}

The work done by Zechen Bai et al\cite{HalluSurvey} is an excellent survey offering a comprehensive, in-depth, and systematic introduction and analysis of the causes, evaluation metrics, and current solutions for hallucinations in Multimodal Large Language Models (MLLMs). According to\cite{HalluSurvey}, hallucinations can originate from data, model, training process and inference process. Our insights highlight a particularly important cause of hallucinations: MLLMs tend to spontaneously ignore visual features. Current MLLM architectures are highly imbalanced as the language model (LLM), with strong priors embedded during massive pretraining, weighs much more than the visual module, suffering significant information loss while extracting visual features. Also when an MLLM generates tokens sequentially in an autoregressive manner during inference, it increasingly focuses on the tokens that have already been generated as the output gets longer and longer, gradually ignoring the input prompt, especially the visual information.\\

Researchers have come up with many solutions for the issue of "Visual Ignorance". LLaVA-1.5\cite{LLAVA_L}, Qwen-vl\cite{Qwen-vl}, Internvl\cite{Internvl} and HallE-Switch\cite{HallE-Switch} have shown that increasing the number of parameters in the vision encoder and improving image resolution can effectively reduce hallucinations. The works in\cite{VisExpert},\cite{Vcoder},\cite{EyesShut} and\cite{VisDetect} enhance the representation of the visual component by integrating visual features extracted from various vision encoders, utilizing visual perception tools such as OCR tools and object detectors, and incorporating perceptual information like depth maps and segmentation masks, allowing the visual part to play a bigger role. To enable MLLM training to benefit from feedback in the same way that LLM training does, Silkie\cite{Silkie}, HA-DPO\cite{HA-DPO}, LLaVA-RLHF\cite{LLaVA-RLHF} and RLHF-V\cite{RLHF-V} leverage feedback from both AI systems(RLAIF) and humans(RLHF) to train a reward model that can identify hallucinations and prefer low-hallucination responses. With regard to inference stage, MARINE\cite{MARINE}, GCD\cite{GCD} and HALC\cite{HALC} adhere to the concept of "guided decoding," utilizing grounded visual objects, grounded visual tokens, and even scores that can accurately measure the degree of hallucination to guide the decoding process of MLLMs. These approaches ensure that the generated language is as visually grounded as possible.

\section{Method}

\subsection{Architecture}  

\begin{figure}[!ht]
\centering  
\vspace{-4mm}
\includegraphics[height=3.5cm]{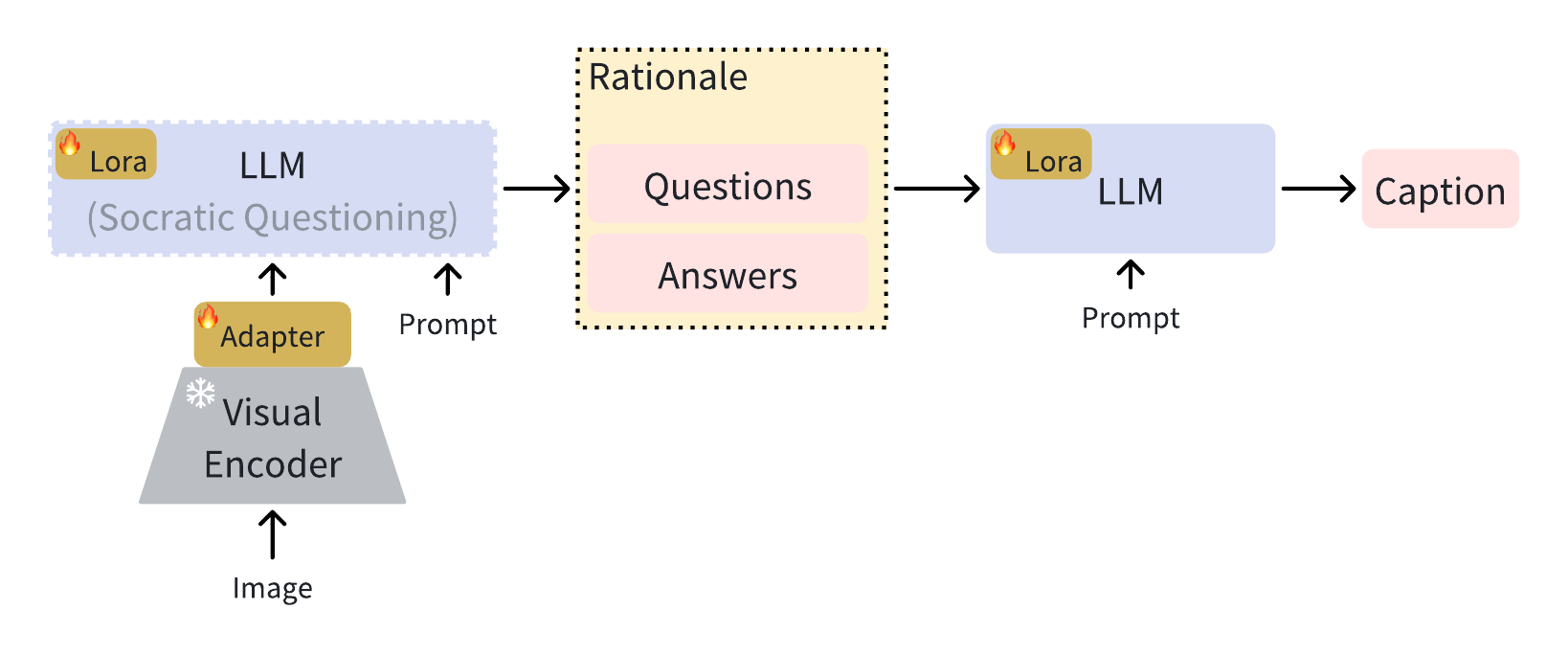}
\vspace{-1mm}
\caption{\shortname{} network architecture. Note that the two LLM modules correspond to a single LLM. The visual encoder outputs visual features that will be mapped by the adapter to visual tokens. The visual tokens, along with the self-ask abd self-answer prompt token, making the LLM generate a rationale comprised of Q\&A pairs. Then the same LLM takes the rationale tokens and the description and summarization prompt tokens to produce the final caption.}
\label{fig:sq_arch}  
  \vspace{-3mm}
\end{figure}

\paragraph{\shortname{} architecture} 
The network archtecture of \shortname{} is illustrated in Figure~\ref{fig:sq_arch}. In order to reduce memory usage, we make the LLM act as  a Question Generator, Question Answer and Visual Summarizer simultaneously. As a Question generator, the LLM generates a list of questions seeking valuable information to help itself correctly interpret the ongoing activity within the given image. As a Question Answerer, the LLM answers these questions one by one (essentially performing VQA tasks) to produce the rationale consisting of the Q\&A pairs. As a Visual Summarizer, the LLM provides final detailed descriptions and summarized captions based on the information encoded in the previous rationale. Note that the dashed LLM module named \textbf{Socratic Questioning} on the left denotes the roles of Question Generator and Question Answerer, while the undashed LLM module on the right denotes the role of Visual Summarizer. The two LLM modules are actually representations of the same LLM. All three functionalities of LLM can be trained jointly, , making the process efficient in terms of memory and computation.

\paragraph{Adapter.} The adapter module can be implemented using a simple linear layer, a multilayer perceptron (MLP), or a cross-attention-based transformer architecture. As the complexity of the network increases, so does the amount of data required for training. It has been demonstrated in LLaVA-1.5 \cite{LLAVA_L} that using a two-layer MLP improves the model's multimodal capabilities compared to using a linear projection. As a tradeoff between computational cost and performance, we have adopted a two-layer MLP as our adapter. It maps vectors from the image feature space to the word embedding space of LLM, aligning the visual features with the textual feature space.

\paragraph{Visual Encoder and Textual Decoder.}
We use a pretrained ViT-L/14~\cite{ViT} as our image encoder and a pretrained Vicuna~\cite{Vicuna} as our LLM. The visual feature and textual embedding spaces are aligned using adapters, which consist of two-layer MLP. 

\section{Generation of the CapQA Dataset}
\label{sec:generation of the dataset}

\subsection{Data Collection}
\label{sec:data_collection}

We collect data from the Consented Activities of People (CAP)~\cite{CAP} dataset, which comprises video clips of daily activities performed by consenting individuals around the world. The CAP dataset contains 1,454,540 clips, categorized into 512 classes of fine-grained activities with labels (like "\emph{person opens car door}") encoding subjects' actions and the objects they are interacting with. The activities are fine-grained because they differ in the subtle details of actions and interacting objects, although they may appear similar overall. We select 20 activities and randomly extract 50 clips from each activity.  From each clip, we chose one key frame that clearly demonstrates the ongoing activity to serve as our final image data with activity label.
\subsection{Designing prompts to automatically generate annotation}

To further annotate the image data obtained in 2.1, we utilize GPT-4v \cite{GPT-4V} to automatically generate the annotations including a list of questions, corresponding answers, a detailed description and a summarized caption. Our meticulously designed prompt for annotations acquisitions is shown at ~\ref{tab:full_example_car_bbox}:

\begin{itemize}[leftmargin=7.5mm]
\setlength{\itemsep}{2pt}
\item 
{\it Questions \& Answers}. To address the ongoing activity depicted in an image, we prompt the GPT-4V model to generate relevant questions and provide corresponding answers. We guide the GPT-4v model to refine its questions and validate that its answers are visually well-founded so that each QA pair is specific, accurate, and meaningful. We also provide the ground truth activity label (excluded in the final produced annotation) such as ``\emph{person opens car door}'' for better alignment of annotations to reality. Please note that the ground truth label will only be used in data generation.\\

\item
{\it Detailed Description}. Based on the information implied in the sequence of produced Q\&A pairs, the GPT-4v model provides a detailed description that includes the person's appearance and actions, the surrounding environment, the attributes and condition of the interacted objects, as well as insights into the person's intentions and potential changes in the situation.\\

\item
{\it Summarized Caption}. Although greatly informative, the detailed description contains lots of redundant information and even some hallucinations, which increases the risk of misleading users. Therefore,  we also prompt the GPT-4v model to condense the detailed description into a summarized caption, a concise expression retaining the core content most relevant to the activity's theme.
\end{itemize}

\begin{table*}[!ht]\centering
\begin{minipage}{1.0\columnwidth}\vspace{0mm}    \centering
\begin{tcolorbox} 
    \centering
    
        \footnotesize
    \begin{tabular}{p{0.97\columnwidth} c}
    \VarSty{ {\bf Prompt} } &\\
    Please come up with 5-8 questions related to the details of the activity and answer them based on the image. If certain questions remain uncertain, further refine those questions, then come up with 5 necessary questions for those uncertain aspects and provide answers. Summarize the refined questions and answers to attempt addressing the uncertain questions again, without exceeding 20 questions in total. Finally, compile all questions and answers to complete two descriptions of the activity depicted in the image. It is known that the activity is '{\emph{person enters car}}', but do not include this phrase in your descriptions. Start with a detailed description, our main task is to detect the activity based on the image, so please provide as detailed a description as possible, related to this main task. You should aim for a granular and comprehensive description of every detail of the activity, within 1000 words; then provide a concise description, simplifying the detailed description to retain only the parts most relevant to the activity, within 400 words. Please self-ask and self-answer again. 
    & \hspace{-3.2cm} \multirow{5}{*}{}\\
    \end{tabular}
\end{tcolorbox}
\vspace{-2mm}
\caption{One example to prompt the gpt-4v model to generate annotations for the given image.}
    \label{tab:full_example_car_bbox}
\end{minipage}
\end{table*}

\subsection{Data Label Format}
\label{sec:data_label_format}

To facilitate future fine-tuning, we have decided to organize the acquired annotations in a multi-round conversation format, similar to that used in LLaVA \cite{LLAVA}. The first round of conversation generates the list of questions and the subsequent rounds consist of Q\&A pairs where the questions are taken from the list in order and answers are given by GPT-4v accordingly. Finally, the last two rounds of conversation elicit the detailed description and summarized caption respectively. An example of annotation formatted into multi-round conversation is shown at Table~\ref{tab:full_example_of_conversation} in the appendix.

We extract key frames from selected activity clips of the \emph{CAP} dataset, leverage GPT-4v to automatically annotating image data and finally organize the annotations into a structured conversation format. This approach enhances the  granularity, accuracy, depth, and comprehensiveness of our annotations, while streamlining the annotation process and optimizing data utility for further analysis.

\subsection{Training}

\subsubsection{Training data format}
\label{sec:training_data_format}

In Section~\ref{sec:data_label_format}, we depict the label format used for our CapQA dataset~\ref{tab:full_example_of_conversation}. 
Assuming a multi-turn conversation $\{\Xmat_{q}^{1}, \Xmat_{a}^{1},\Xmat_{q}^{2}, \Xmat_{a}^{2},...,\Xmat_{q}^{T}, \Xmat_{a}^{T}\}$ consists of $T$ turns, we denote the human's question in the $j$-th turn of conversation as $\Xmat_{q}^{j}$ and system(like GPT)'s answer in the $j$-th turn of conversation as $\Xmat_{a}^{j}$.
\paragraph{Questions Generation.} The first turn $[\Xmat_{q}^{1}, \Xmat_{a}^{1}]$ is specifically designed to train the LLM to function as a Question Generator. $\Xmat_{q}^{1}$ denotes a carefully crafted prompt requesting the questions generation, while $\Xmat_{a}^{1}$ denotes the list of questions generated upon $\Xmat_{q}^{1}$. Again, the questions would guide the LLM to capture fine-grained details of human activity so as to correctly interpret the image.

\paragraph{Answers Generation.} 
$(\Xmat_{q}^{2}, \Xmat_{q}^{3},...,\Xmat_{q}^{T-3}, \Xmat_{q}^{T-2})$ are the individual questions contained in the list $\Xmat_{a}^{1}$, while $(\Xmat_{a}^{2}, \Xmat_{a}^{3},...,\Xmat_{a}^{T-3}, \Xmat_{a}^{T-2})$ are their corresponding answers. Thus, the turns $(\Xmat_{q}^{2}, \Xmat_{a}^{2},...,\Xmat_{q}^{T-2}, \Xmat_{a}^{T-2})$ of the conversation are well-suited to train the LLM to function as a Question Answering and finish the VQA tasks well.
\paragraph{Detailed Description Generation ($[\Xmat_{q}^{T-1},\Xmat_{a}^{T-1}]$).} In step $T-1$,  the system is instructed to generate a detailed description that thoroughly articulates the contents of the image, including objects within the scene, the background, and the attributes and actions of people. The goal is to capture as much detailed information as possible to enhance the depth and comprehensiveness of the description, thereby laying a solid foundation for future reasoning.

\paragraph{Summarized Caption Generation ($[\Xmat_{q}^{T},\Xmat_{a}^{T}]$).} Following that, in step $T$, the system is instructed to generate a condensed caption. This step distills the information in the detailed description by focusing on the core elements. It includes only the most significant features and actions from the image, aiming to offer a clear, succinct, and informative caption without overwhelming complexity.

\begin{figure}[h!]
\centering  
\vspace{-4mm}
\includegraphics[height=3.2cm]{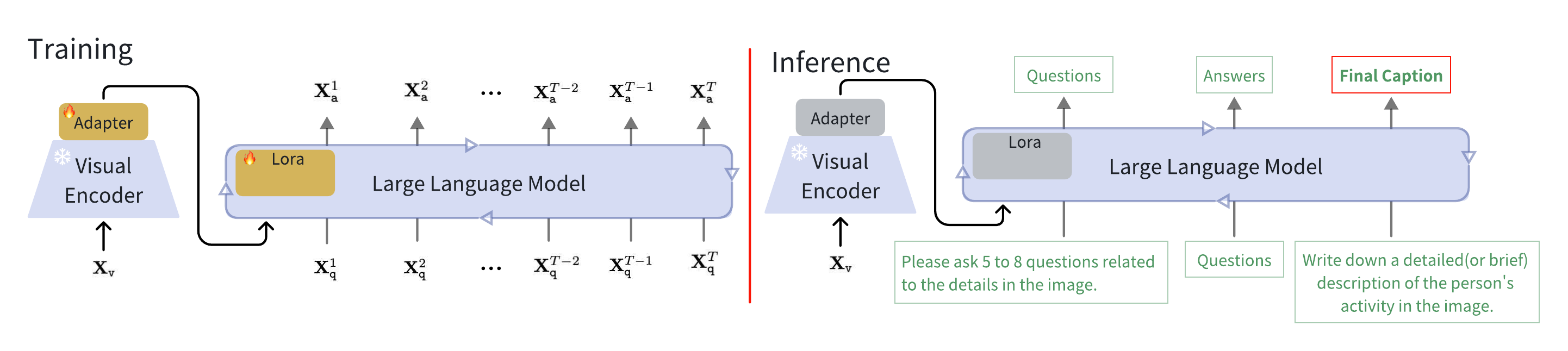}
\vspace{-1mm}
\caption{Illustrations of the training (left) and 3-turn inference (right) processes of \shortname{}.}
\label{fig:sq_inference}  
    \vspace{-3mm}
\end{figure}

\subsubsection{Training Procedure}
We train our model using a classical two-stage process: first pretraining, followed by instruction-tuning.
\paragraph{Stage 1: Pretrain} 
We utilize the LLaVA-CC3M-Pretrain-595K dataset \cite{LLAVA},comprised of 595K image-text pairs filtered from CC3M, to pretrain the adapter of \shortname{}. The purpose of this stage is to achieve a good alignment between the visual feature space and the token embedding space of LLM. The parameters of both the image encoder and the LLM are frozen throughout the pretraining phase.

\paragraph{Stage 2. Instruction-Tune} We fine-tune our \shortname{} model using 666K image-text pairs. It contains llava\_v1\_5\_mix665k~\cite{LLAVA_L} and CapQA\_{0.9k} dataset introduced elaborately in Sections ~\ref{sec:data_label_format} and ~\ref{sec:training_data_format}. We processed the questions from the Conv58k dataset~\cite{LLAVA_L}, included in the llava\_v1\_5\_mix665k~\cite{LLAVA_L}, these questions were reorganized to conform to the data format described in Section~\ref{sec:data_label_format}. To prevent overfitting, during training, we randomly insert the generated question list at any round of the conversation.
During this phase, the image encoder remains frozen, while the adapters and LLM (using LoRA \cite{LoRA}) are fine-tuned. We perform instruction-tuning of the LLM on the prediction tokens, using the same auto-regressive training objective as LLaVa \cite{LLAVA}:
\begin{equation}
    p( \Xmat_{\texttt{a}} |  \Xmat_{\texttt{v}}, \Xmat_{\texttt{q}}) =
    \prod_{i=1}^{L} p_{\thetav} (  {\color{mygreen} \xv_i}
| \Xmat_{\texttt{v}}, \Xmat_{\texttt{q}, <i}, \Xmat_{\texttt{a}, <i}),
    \label{eq:auto_regressive}
\end{equation}

Where $L$ is the token sequence length, $\Xmat_{\texttt{v}}$ stands for the visual input (visual tokens), $\Xmat_{\texttt{q}}$ and $\Xmat_{\texttt{a}}$ stand for the tokens of human instructions and system answers, respectively, across all $T$ rounds of conversation. $\Xmat_{\texttt{q}, <i}$ and $\Xmat_{\texttt{a}, <i}$ are respectively the human instructions tokens and system answers tokens in all turns before the currently predicted token ${\color{mygreen} \xv_i}$. 

\subsection{Inference}

 The inference process is illustrated in the right side of Figure~\ref{fig:sq_inference}. We can choose to employ 1-turn or 3-turn inference. Simply, 1-turn inference directly produces the final caption based on the given image, problem statement and context(\textbf{Inputs}). As the right side of Figure~\ref{fig:sq_inference} shows, 3-turn inference first prompt the LLM to generate a list of questions based on the \textbf{Inputs}, then make the LLM provide visually grounded answers of these questions and finally let the LLM generate the detailed description and summarized caption, where the more concise caption is treated as the final output. Experiments show that 1-turn inference is better suited for straightforward problems while 3-turn inference works better for complicated problems requiring multi-step reasoning and fine-grained details.

\section{Experiments}

\subsection{CapQA}

\begin{wrapfigure}{r}{0.5\textwidth}
  \begin{minipage}{0.5\textwidth}
\centering  
\vspace{-4mm}
\scalebox{0.88}{
\begin{tabular}{l|l l}
\toprule
Method & HalS & QQS \\
\midrule
InstructBLIP & 87.4 & 78.4 \\
LLaVA-1.5 & 69.3 & 31.5 \\
LLaVA-1.5 + $SQ_{train}$ & 90.9 & 92.3 \\
LLaVA-1.5 + $SQ_{train}$ +  \text{3turns} & 93.0 & - \\
\bottomrule
\end{tabular}
}
\vspace{2mm}
\captionof{table}{Ablation w/o multi-turn train/inference on CapQA. We adapt vicuna7b as LLM. HalS: Hallucination Score; QQS: Questions Quality Score. Evaluation are GPT4~\cite{GPT4}-aid.}  
\label{tab:capqa_result}  
  \end{minipage}
\end{wrapfigure}

CapQA, proposed in this paper, is a novel mini-dataset consisting of 982 images, each associated with a multi-turn conversations. The dataset is divided into a training set and a test set, with the training set containing 882 samples and 11.9k QA pairs, and the test set containing 100 samples and 1.4k QA pairs. 

We designed two evaluation metrics: (A). Hallucination, measuring the degree of hallucination in detailed descriptions, with higher scores indicating less hallucination. (B). Questions Quality, reflecting model's ability to generate questions, with higher scores reflecting better quality, diversity, and effectiveness. 
The calculation method for the score can be expressed as follows:

\begin{equation}
    \text{HalS} = \frac{\text{HalS}_{\text{pred}}}{\text{HalS}_{\text{gt}}} \text{; } \\
    \text{QQS} = \frac{\text{QQS}_{\text{pred}}}{\text{QQS}_{\text{gt}}}
    \label{eq:hal_score_compute}
\end{equation}

In Eq~\ref{eq:hal_score_compute}, \(\text{HalS}_{\text{pred}}\) represents the average score of all model predictions reviewed by GPT-4~\cite{GPT4}, while \(\text{HalS}_{\text{gt}}\) denotes the average score of all labels reviewed by GPT-4~\cite{GPT4}. Similarly, \(\text{QQS}_{\text{pred}}\) is the average score of all model predictions reviewed by GPT-4~\cite{GPT4}, and \(\text{QQS}_{\text{gt}}\) is the average score of all labels reviewed by GPT-4~\cite{GPT4}. The prompt used to instruct GPT-4 for scoring is shown in Table~\ref{tab:prompt_for_gpt4_eval}.

As shown in Table~\ref{tab:capqa_result}, our proposed \shortname{} framework leads to a 31.2\% improvement in the hallucination score and a significant increase in the question quality score from 31.5 to 92.3. Additionally, employing a 3-turn inference mode, which includes question-answer-caption 3 steps during the inference phase, further reduces hallucination by 2.3\%. Without increasing computational cost, our SQ method effectively reduces the model's hallucination while generating detailed descriptions.

\subsection{POPE}

POPE~\cite{POPE} is focused on assessing the hallucinations in MLLMs by testing if the MLLMs can correctly tell the existence of objects in images. It employs different sampling methods to construct negative samples, including random, popular, and adversarial sampling. In the random sampling setting, objects that are not present in the image are chosen randomly. For the popular setting, the absent objects are selected from a pool of the most frequently occurring objects. In the adversarial setting, objects that commonly co-occur but are not present in the image are used as negative samples.
We achieved better performance than Woodpecker~\cite{Woodpecker} on the POPE benchmark and attained state-of-the-art (SOTA) F1 scores across three different modes.

\begin{table*}[!ht]
\small
\centering
\aboverulesep = 0.2mm 
\resizebox{0.76\textwidth}{!}{%
\begin{tabular}{llccc|c|c}
\toprule
Setting                      & Method                    & Accuracy       & Precision      & Recall         & F1-Score       & Yes Rate \\ 
\midrule
\multirow{8}{*}{\textit{Random}}      & {LLaVA~\cite{LLAVA}}    & 86.00          & 87.50          & 84.00          & 85.71    & 48.00        \\  
                             & {LLaVA + Woodpecker~\cite{Woodpecker}}     & 87.67 & 95.93 & 78.67          & 86.45 & 41.00        \\  
                             & {MiniGPT-4~\cite{MiniGPT-4}}       & 54.67          & 57.78          & 34.67          & 43.33          & 30.00        \\  
                             & {MiniGPT-4 + Woodpecker}       & 85.33          & 92.06          & 77.33          & 84.06          & 42.00        \\  
                             & {mPLUG-Owl~\cite{mPLUG-Owl}}   & 62.00          & 57.26          & \textbf{94.67} & 71.36          & 82.67        \\  
                             & {mPLUG-Owl + Woodpecker}  & 86.33          & 93.60          & 78.00          & 85.09          & 41.67        \\  
                             & {Otter~\cite{Otter}}        & 72.33          & 66.18          & 91.33    & 76.75          & 69.00        \\  
                             & {Otter + Woodpecker}      & 86.67    & 93.65    & 78.67          & 85.51          & 42.00        \\ 
                             & {LLaVA-1.5~\cite{LLAVA_L}} & 88.18 & \textbf{97.45} & 79.13 & 87.34 & 41.85 \\ 
                             & \textbf{Ours}      & \textbf{89.21}    & 95.69    & 82.80          & \textbf{88.78}          & 44.60        \\ 
                             \midrule
\multirow{8}{*}{\textit{Popular}}     & {LLaVA~\cite{LLAVA}}     & 76.67          & 72.22          & 86.67          & 78.79          & 60.00        \\  
                             & {LLaVA + Woodpecker}    & 80.67          & 83.82          & 76.00          & 79.72          & 45.33        \\  
                             & {MiniGPT-4~\cite{MiniGPT-4}} & 56.67          & 58.77          & 44.67          & 50.76          & 38.00        \\  
                             &  {MiniGPT-4 + Woodpecker} & 82.33          & 85.40    & 78.00          & 81.53          & 45.67        \\  
                             & {mPLUG-Owl~\cite{mPLUG-Owl}} & 57.33          & 54.20          & \textbf{94.67} & 68.93          & 87.33        \\  
                             & {mPLUG-Owl + Woodpecker} & 83.00    & 84.14          & 81.33          & 82.71    & 48.33        \\  
                             & {Otter~\cite{Otter}}     & 67.33          & 61.71          & 91.33    & 73.66          & 74.00        \\  
                             & {Otter + Woodpecker}   & 84.33 & 88.15 & 79.33          & 83.51 & 45.00        \\ 
                             & {LLaVA-1.5~\cite{LLAVA_L}} & 87.27 & \textbf{94.51} & 79.13 & 86.14 & 41.87 \\ 
                             & \textbf{Ours}      & \textbf{87.53}    & 91.46    & 82.80          & \textbf{86.91}          & 45.27        \\ 
                             \midrule
\multirow{8}{*}{\textit{Adversarial}} & {LLaVA~\cite{LLAVA}}     & 73.33          & 69.02          & 84.67          & 76.05          & 61.33        \\  
                             & {LLaVA + Woodpecker} & 80.67          & 82.86          & 77.33          & 80.00          & 46.67        \\  
                             & {MiniGPT-4~\cite{MiniGPT-4}} & 55.00          & 56.88          & 41.33          & 47.88          & 36.33        \\  
                             & {MiniGPT-4 + Woodpecker} & 82.33    & 83.92    & 80.00          & 81.91    & 47.67        \\  
                             & {mPLUG-Owl~\cite{mPLUG-Owl}} & 56.33          & 53.51          & \textbf{96.67} & 68.88          & 90.33        \\  
                             & {mPLUG-Owl + Woodpecker} & 81.00          & 82.07          & 79.33          & 80.68          & 48.33        \\  
                             & {Otter~\cite{Otter}}     & 66.67          & 61.16          & 91.33    & 73.26          & 74.67        \\  
                             & {Otter + Woodpecker}  & 83.00 & 85.61 & 79.33          & 82.35 & 46.33        \\ 
                             & {LLaVA-1.5~\cite{LLAVA_L}} & 85.13 & \textbf{89.92} & 79.13 & 84.18 & 44.00 \\ 
                             & \textbf{Ours}      & \textbf{85.23}    & 87.03    & 82.80          & \textbf{84.87}          & 47.57        \\ 
                             \bottomrule
\end{tabular}%
}
\caption{Results on POPE~\cite{POPE}. The best performances within each setting are \textbf{bolded}.}
\label{tab:pope}
\end{table*}

\subsection{Comparative Experiment}

The experimental results presented in Table~\ref{tab:contrast_experiment} demonstrate the superiority of our proposed method compared to several state-of-the-art (SoTA) methods across six benchmarks. Our method utilizes the Vicuna-7B~\cite{Vicuna} large language model with 336 resolution, 558K pre-train data, and 666K(llava\_v1\_5\_mix665k~\cite{LLAVA_L} + CapQA\_0.9k) fine-tune data. It achieves a Hallucination Rate (HalR) of 0.57 and an MMHal Average Score (AvgS) of 2.16, outperforming other methods in these metrics. Notably, our method also excels in the LLaVA-QA90~\cite{LLAVA} benchmark with a score of 81.3, and shows competitive performance in the LLaVA-Bench (In-the-Wild)~\cite{LLAVA}, MME~\cite{MME}, ScienceQA-IMG~\cite{ScienceQA}, and TextVQA~\cite{TextVQA} benchmarks with scores of 66.8, 1523.4, 68.37, and 58.57, respectively. These results highlight the robustness of our approach in reducing hallucinations and improving overall question quality without additional computational load. The introduction of all datasets has been moved to \ref{sec:datasets} in the appendix.

\begin{table}[t!]
\centering
\scalebox{0.74}{
\begin{tabular}{l|llll|ll|lllll}
\toprule
\multirow{2}{*}{Method}  & \multirow{2}{*}{LLM}  & \multirow{2}{*}{Res} & \multirow{2}{*}{PT} & \multirow{2}{*}{IT} & \multicolumn{2}{c|}{MMHal} & \multirow{2}{*}{LLaVA$^\text{qa90}$} & \multirow{2}{*}{LLaVA$^\text{W}$} & \multirow{2}{*}{MME} & \multirow{2}{*}{SQA$^\text{I}$} & \multirow{2}{*}{VQA$^\text{T}$} \\
 &  &  &   &  & AvgS & HalR & & & \\
\midrule
BLIP-2~\cite{BLIP2} & Vicuna-13B & 224 & 129M & - & - & - & - & 38.1 & 1293.8 & 61.0 & 42.5 \\
InstructBLIP~\cite{InstructBLIP} & Vicuna-7B & 224 & 129M & 1.2M & 2.1 & 0.58 & \textbf{85.8} & 60.9 & - & 60.5 & 50.1 \\
InstructBLIP~\cite{InstructBLIP} & Vicuna-13B & 224 & 129M & 1.2M & 2.14 & 0.58 & - & 58.2 & 1212.8 & 63.1 & 50.7 \\
Qwen-VL~\cite{Qwen-vl} & Vicuna-7B & 448 & 1.4B & 50M & - & - & - & - & - & 67.1 & \textbf{63.8} \\
Qwen-VL-Chat~\cite{Qwen-vl} & Vicuna-7B & 448 & 1.4B & 50M & - & - & - & - & 1487.5 & 68.2 & 61.5 \\
LLaVA-1.5~\cite{LLAVA_L} & Vicuna-7B & 336 & 558K & 665K & 2.04 & 0.61 & 79.9 & 63.4 & 1510.7 & 66.8 & 58.2 \\
\midrule
\textbf{Ours} & Vicuna-7B & 336 & 558K & 666K & \textbf{2.16} & \textbf{0.57} & \underline{81.3} & \textbf{66.8} & \textbf{1523.4} & \textbf{68.37} & 58.57 \\
\bottomrule
\end{tabular}
}
\vspace{1mm}
\caption{Comparison with SoTA methods on 6 benchmarks. MMHal: MMHal-Bench~\cite{MMHal}, AvgS: Average Score; HalR: Hallucination Rate; ; LLaVA$^\text{qa90}$: LLaVA-QA90~\cite{LLAVA}; LLaVA$^\text{W}$: LLaVA-Bench (In-the-Wild)~\cite{LLAVA}; SQA$^\text{I}$: ScienceQA-IMG~\cite{ScienceQA}(zero-shot); VQA$^\text{T}$: TextVQA~\cite{TextVQA}; MME~\cite{MME}.}
\vspace{-3mm}
\label{tab:contrast_experiment}
\end{table}

\section{Conclusion}

In this work, we introduce the Socratic Questioning (SQ), an flexible, reliable and effective framework for visual reasoning and question answering that fits well to lightweight Multimodal Large Language Models (MLLMs). SQ combines Chain of Thought (CoT) reasoning and visual instruction tuning through heuristic self-questioning, effectively reducing hallucinations and training costs while improving fine-grained visual detail description and zero-shot reasoning. Our experiments, including those with the new CapQA dataset, demonstrate SQ's effectiveness in reducing hallucinations and improving visual description quality. By efficiently utilizing lightweight MLLMs, SQ provides a cost-effective, high-performance solution for complex visual tasks, paving the way for future research in multimodal reasoning.

\textbf{Discussion.} This work is merely an exploration of heuristic self-questioning, and there are areas that require further improvement. For example, designing a reasonable loss function to constrain the model to ask more effective questions that benefit the overall task, and enhancing fine-grained visual information using Visual large Model(VLM) encoders with region alignment capabilities (such as GLIP~\cite{GLIP}, SAM~\cite{SAM}, etc.). These aspects are left for future long-term research.

\textbf{Acknowledgements.} 
We thank the LLaVA team for their awesome work at exploration of MLLMs.
We thank the LLaMA team for giving us access to their models, and open-source projects, including Alpaca and Vicuna.

\bibliographystyle{plain}
\bibliography{references}

\clearpage
\appendix

\section{Training Parameters Detail}
\paragraph{Pre-training} 

We directly use the pretrained weights of LLaVA-1.5~\cite{LLAVA_L}. You can download from: \href{https://github.com/haotian-liu/LLaVA/blob/main/docs/MODEL_ZOO.md}{https://github.com/haotian-liu/LLaVA/blob/main/docs/MODEL\_ZOO.md}

\paragraph{Instruct Fine-tuning} 

We conduct instruction fine-tuning training of our model on four NVIDIA A800-SXM4-80GB GPUs, which takes approximately 28 hours. The hyperparameters are shown in Table~\ref{tab:hyperparameter}.

\begin{table}[!ht]
\centering
\scalebox{0.76}{
\begin{tabular}{l| c c}
\toprule
Hyperparameter & Finetune \\
\midrule
batch size & 128 \\
lr & 2e-5 \\
lr schedule & cosine decay \\
lr warmup ratio & 0.03 \\
weight decay & 0 \\
epoch & 1 \\
optimizer & AdamW \\
DeepSpeed stage & 3 \\
\bottomrule
\end{tabular}
}
\caption{
\textbf{Hyperparameters} of Fine-tuning of \shortname{}.
}
\label{tab:hyperparameter}
\end{table}

\section{Prompt}

\begin{table*}[!ht]\centering

\begin{minipage}{0.99\columnwidth}\vspace{0mm}    \centering
\begin{tcolorbox} 
    \centering
    \small
     \hspace{-6mm}
      \scriptsize
    \begin{tabular}{p{0.99\columnwidth}}
    \VarSty{ {\bf Hallucination} } \\

We would like to request your feedback on the performance of two AI assistants in response to the user question displayed above. The user asks the question on observing an image. For your reference, the visual content in the image is represented with a few sentences describing the image.\\nPlease rate their responses based on the hallucination (i.e., unreal or unfounded content). Each assistant receives an overall score on a scale of 1 to 10, where a lower score indicates fewer hallucinations and better performance. Please first output a single line containing only two values indicating the scores for Assistant 1 and Assistant 2, respectively. The two scores are separated by a space. In the subsequent line, please provide a comprehensive explanation of your evaluation, avoiding any potential bias and ensuring that the order in which the responses were presented does not affect your judgment. \\
\\
    \VarSty{ {\bf Questions Quality} } \\

We would like to request your feedback on the performance of two AI assistants in generating questions based on the image content. The task for the assistant is to propose several diverse and effective questions, aimed at obtaining a more accurate detailed description. For your reference, we will provide additional information about the image and questions (such as the expected questions, human-generated questions, and hints given by annotators). Note that the assistant can only see the image content and question text, and all other reference information is used to help you better understand the questions and content of the image only. The major criteria for evaluation are the diversity, effectiveness, and accuracy of the questions generated.\\nEach assistant receives an overall score on a scale of 1 to 5, where a higher score indicates better overall performance. Please first output a single line containing only two values indicating the scores for Assistant 1 and Assistant 2, respectively. The two scores are separated by a space. In the subsequent line, please provide a comprehensive explanation of your evaluation, avoiding any potential bias and ensuring that the order in which the responses were presented does not affect your judgment. \\
\end{tabular}

\end{tcolorbox}
    
\vspace{-2mm}
\caption{Prompt used in CapQA GPT4-aid Evaluation.}
    \label{tab:prompt_for_gpt4_eval}
\end{minipage}
\end{table*}

\section{Datasets}
\label{sec:datasets}

\subsection{MMHal}
MMHal is comprised of 96 delicately designed image-question pairs, ranging in 8 question
categories × 12 object topics. MMHal concentrates on detecting hallucinations within the LMM responses and adopts general, realistic, and open-ended questions to better reflect the response quality in real-world user-LMM interactions. The images are from the validation and test sets of OpenImage to avoid data leakage. The questions, asking LLM to figure out the object attributes, spatial relations, make counting, provide holistic description and etc, are created in an adversarial manner to make LLM hallucinates on purpose. As a result, MMHal offers a great assessment on LLM's capability to robustly resist various kinds of hallucinations.\\

\subsection{MME} 
MME, introduced in \cite{MME}, is a comprehensive MLLM Evaluation benchmark consisting of 1k - 2k images and instruction-answer pairs. MME has four main characters:\\

1. MME offers a comprehesive assessment for different aspects of a MLLM's ability including perception (coarse-grained and fine-grained object recognition and OCR) and cognition (commonsense reasoning, numerical calculation, text translation, and code reasoning), up to totally 14 subtasks.\\

2. All instruction-answer pairs are manually constructed and great proportion of images are newly collected in order to avoid data leakage.\\

3. The instructions are made concise so as to be similar to commonly used ones. The unfair advantage of prompt engineering is avoided. \\

4. The answers are simple "yes" or "no", which is accurate, objective and convenient for quantitative analysis.\\

Hence, MME is an accurate, objective, fair and comprehensive benchmark for MLLM's visual perception and cognition capabilities. 
\subsection{TextVQA}
TextVQA dataset, introduced in \cite{TextVQA}, contains 28408 images on which 45336 questions are asked by human annotators. The images, selected from the Open Images dataset, belong to the categories that tend to contain text e.g. “billboard”, “traffic sign”, “whiteboard”. The questions require reading and reasoning about text in the image. Date are organized in the format of question-image pairs where each has 10 ground truth answers provided by humans. This benchmark evaluate model's reasoning ability specialized in Optical Character Recognition (OCR). Our \shortname{} achieve state-of-art performance without a specialized OCR module.
\subsection{LLaVA-Bench(In-the-Wild)}
LLaVA-Bench(In-the-Wild) is introduced in the work of LLaVa\cite{LLAVA} created to evaluate models ability to handle challenging tasks and generalize to new domain. It has totally 24 images, each comes with a manually annotated detailed description, of indoor and outdoor scenes, memes, paintings, sketches, etc. Authors also provide a list of 60 questions, from which individual questions are properly selected to be associated with each image. In this way, LLaVA-Bench(In-the-Wild) works well as a benchmark for visual captioning and question answering that require strong spatial awareness and background knowledge.
\subsection{LLaVA-QA90}
LLaVA-QA90 is also introduced in the paper of LLaVa\cite{LLAVA}. Authors select 90 images from COCO-Val-2014 and leverage the data generation pipeline introduced in the paper to annotate them. As a result, each image is associated with a detailed description, a multi-round conversation and a complex reasoning Q\&A pair. Thus, LLaVA-QA90 serves as a benchmark for evaluating model's capability to conduct a long conversation, make a detailed description and solve a complex reasoning problem based on an image.

\section{Examples}

\begin{figure}[!t]
\centering  
\vspace{-4mm}
\includegraphics[height=5.5cm]{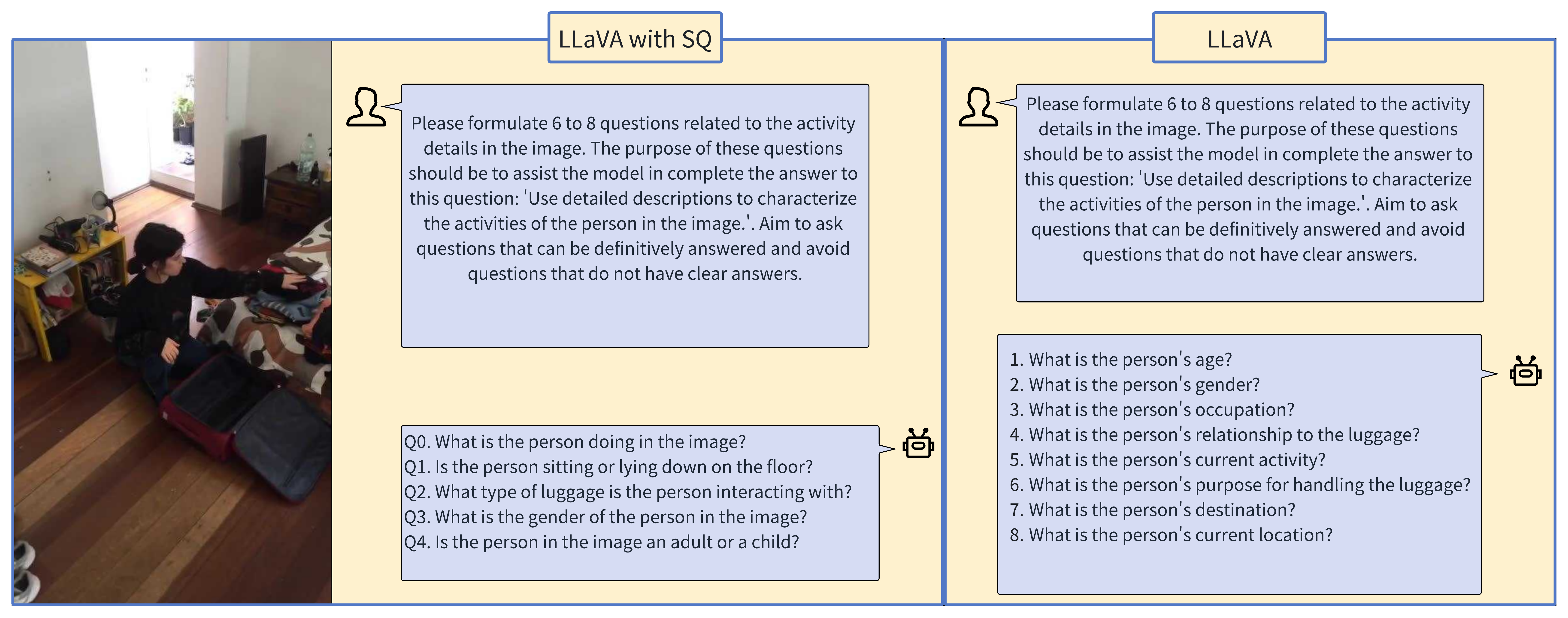}
\vspace{-1mm}
\caption{Comparison of Questions Generation on LLaVA-with-SQ and LLaVA.}
\label{fig:question_gen_constrast}  
    \vspace{-3mm}
\end{figure}

\begin{figure}[!ht]
\centering  
\vspace{-4mm}
\includegraphics[height=5.5cm]{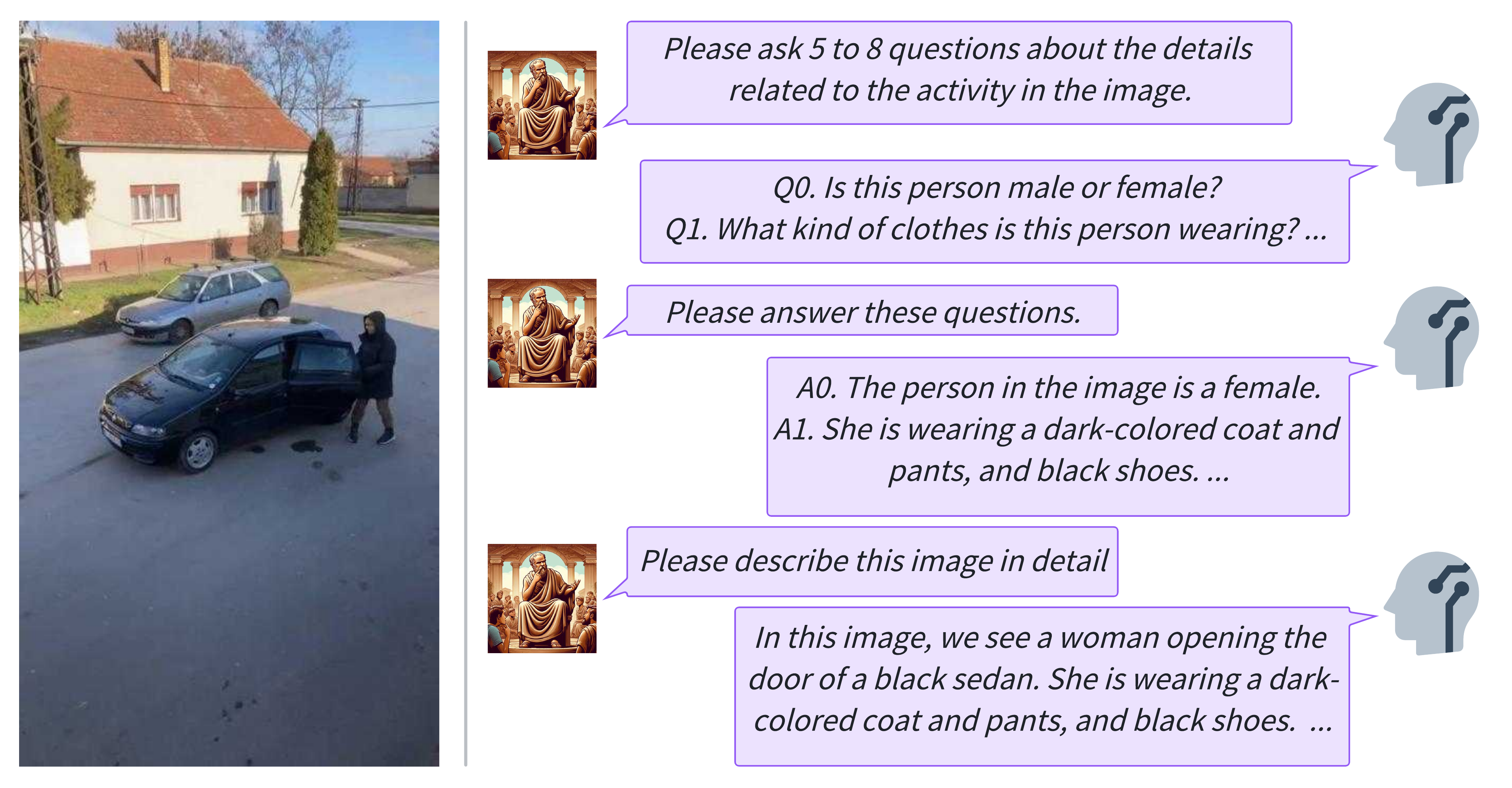}
\vspace{-1mm}
\caption{An example of \shortname{}. \shortname{} starts by asking questions about the given image for heuristic self-questioning. Then it tries to give visually grounded answer to those questions, essentially performing VQA tasks. Finally, \shortname{} summarizes the information encoded in the Q\& A to produce informative and insightful detailed description.}
\label{fig:sq_example}  
  \vspace{-3mm}
\end{figure}


The conversation in Table~\ref{tab:example_cap_conv} is based on Figure~\ref{fig:sq_example}.

\begin{table*}[!ht]\centering
\begin{minipage}{0.99\columnwidth}\vspace{0mm}    \centering
\begin{tcolorbox} 
    \centering
   
     \hspace{-4mm}
      \scriptsize
    \begin{tabular}{p{0.99\columnwidth}}
   \VarSty{ {\bf Conversation in 3-turn mode} } \\
Question:  \\
Please formulate 6 to 8 questions related to the activity details in the image. The purpose of these questions should be to assist the model in complete the answer to this question: 'Use detailed descriptions to characterize the activities of the person in the image.'. Aim to ask questions that can be definitively answered and avoid questions that do not have clear answers. \\
=== \\
Answer: \\
Q0. What is the person doing in the image?
Q1. What type of vehicle is the person getting out of?
Q2. Is the person getting out of a car or a truck?
Q3. What is the person wearing while getting out of the vehicle?
Q4. Is the person standing on a sidewalk or in the street?
Q5. What can be inferred about the person's intentions or actions from the image? \\
=== \\
Question: \\ 
Please answer all the questions one by one directly, answers are split by line break.\\
=== \\
Answer: \\
Q0. What is the person doing in the image?
A. The person is getting out of a car, specifically a black sedan.
Q1. What type of vehicle is the person getting out of?
A. The person is getting out of a black sedan.
Q2. Is the person getting out of a car or a truck?
A. The person is getting out of a car, not a truck.
Q3. What is the person wearing while getting out of the vehicle?
A. The person is wearing a hooded sweatshirt while getting out of the car.
Q4. Is the person standing on a sidewalk or in the street?
A. The person is standing in the street, not on a sidewalk.
Q5. What can be inferred about the person's intentions or actions from the image?
A. From the image, it can be inferred that the person is either arriving at or departing from their destination, as they are getting out of the car. The fact that they are wearing a hooded sweatshirt suggests that the weather might be cold or that they are prepared for the possibility of cold weather during their journey. \\
=== \\
Question: \\
Use detailed descriptions to characterize the activities of the person in the image. \\
=== \\
Answer: \\
In the image, a person is getting out of a black sedan parked on the street. The individual is wearing a hooded sweatshirt, which indicates that the weather might be cold or that they are prepared for the possibility of cold weather during their journey. The person is standing in the street, not on a sidewalk, which could suggest that they are either arriving at or departing from their destination. The presence of a second car parked nearby further emphasizes the idea that this might be a parking spot or a location where multiple people are getting in and out of their vehicles. The person's actions in the image are typical for someone who is either arriving at or departing from a destination, and their choice of clothing reflects the potential weather conditions they might be facing. \\
=== \\

\VarSty{ {\bf Conversation in 1-turn mode} } \\
Question: \\
Use detailed descriptions to characterize the activities of the person in the image. \\
=== \\
Answer: \\
In the image, a person is walking towards a black car that is parked in a driveway. The car is positioned near the curb, and the person is likely getting ready to enter the vehicle. The person is wearing a black jacket, which suggests that the weather might be cool or the person is dressed for a specific occasion. The car is a small, black sedan, and \textcolor{red}{there are two other cars visible in the background}, one of which is parked further away from the main car.  \\
    \end{tabular}
\end{tcolorbox}
\vspace{-2mm}
\caption{An example of a conversation in 1-turn mode and 3-turn mode, where the hallucinated parts are highlighted in red.}
    \label{tab:example_cap_conv}
\end{minipage}
\end{table*}

\clearpage
\section{Label data format of CapQA}

\begin{table*}[!ht]\centering
    \begin{minipage}{1.0\columnwidth}\vspace{0mm}    \centering
    \begin{tcolorbox} 
        \centering
        
            \footnotesize
        \begin{tabular}{p{0.97\columnwidth} c}
        \VarSty{ {\bf Conversation} } &\\
        \texttt{\bf{Human:}} Please formulate 5 to 8 questions related to the activity details in the image. If some questions are uncertain, further refine them, and pose an additional 5 questions specifically targeting these uncertainties, with a total not exceeding 20 questions. The purpose of these questions should be to assist the model in determining the type of activity and in uncovering details of the activity. Aim to ask questions that can be definitively answered and avoid questions that do not have clear answers.\par<image> \\
        
        \texttt{\bf{GPT:}} Q0. What specific part of the vehicle is the person located at?\par 
        Q1. Has the person opened the vehicle door?\par
        Q2. What time of day is the activity taking place?\par
        Q3. Does the person's expression or body language indicate they are in a hurry or relaxed?\par
        Q4. Is the surrounding environment quiet?\par
        Q5. What type and condition is the vehicle in?\par
        Q6. Is the person carrying any items?\par
        Q7. Does the person appear to be preparing to drive away, or have they just arrived?\par
        Q8. Is the person checking inside the vehicle or the surrounding area?\par
        Q9. Is the vehicle door fully open, or only partially open?\par
        Q10. Are the person's hands visible, and are they performing any specific actions?\par
        Q11. Does the person's standing posture suggest they are about to enter the vehicle?\par
        Q12. Besides the vehicle and person, are there any other significant objects or actions?

        \texttt{\bf{Human:}} What specific part of the vehicle is the person located at?

        \texttt{\bf{GPT:}} The person is standing by the side of the vehicle, near the driver's seat door.

        \texttt{\bf{Human:}} Has the person opened the vehicle door?

        \texttt{\bf{GPT:}} Yes, the vehicle door is open.

        \texttt{\bf{Human:}} What time of day is the activity taking place?

        \texttt{\bf{GPT:}} Considering the surrounding light, it appears to be nighttime.

        \texttt{\bf{Human:}} Does the person's expression or body language indicate they are in a hurry or relaxed?

        \texttt{\bf{GPT:}} It's hard to discern the person's expression due to the brightness of the photo, but the body language seems relaxed.

        \texttt{\bf{Human:}} Is the surrounding environment quiet?

        \texttt{\bf{GPT:}} The environment appears to be quiet, with no noticeable activity from other people or vehicles.

        \texttt{\bf{Human:}} What type and condition is the vehicle in?

        \texttt{\bf{GPT:}} The vehicle is a black sedan with a somewhat reflective surface, appearing to be in good condition.

        \dots















        \texttt{\bf{Human:}} Write down a detailed description of the person's activity in the image.

        \texttt{\bf{GPT:}} Detailed Description: \par
        In this nighttime photo, a person is near a black sedan parked at the side of a street. The streetlights cast a soft glow, and the vehicle’s door is fully open. The person stands at the driver’s side door, leaning towards the inside of the car, apparently observing the interior or preparing to enter. They are dressed in dark clothing, blending into the night. The surrounding environment is very quiet, with no other pedestrians or vehicles, reflecting a tranquil atmosphere. The person appears not to be carrying any items, suggesting they might be near home and do not need to carry extra items.

        \texttt{\bf{Human:}} Summarize the details of the person's activity in the image.

        \texttt{\bf{GPT:}} Concise Description: \par
        At night, a person dressed in dark clothing is preparing to enter a black sedan parked on the street side. The door is fully open, and they seem to be getting ready to sit in the driver’s seat. The environment is quiet, with no other apparent activities, giving the whole scene a peaceful nighttime ambiance.

    & \hspace{-3.2cm} \multirow{5}{*}{}
    \\
        \end{tabular}
    \end{tcolorbox}
    \vspace{-2mm}
    \caption{An example of a conversation labeling format.}
        \label{tab:full_example_of_conversation}
    \end{minipage}
    \end{table*}


\begin{table}[t!]
\centering
\scalebox{0.74}{
\begin{tabular}{l|l|l}
\toprule
\multirow{2}{*}{Method}  & \multirow{2}{*}{VCR}  & \multirow{2}{*}{SNLI-VE} \\
 & & \\
\midrule
LLaVA-13b-v0 & 28.3 & 40.3 \\
\midrule
LLaVA-7b-v1.5 & 25.06 & 0.06 \\
\midrule
LLaVA-13b-v1.5 & 25.18 & 45.78 \\
\midrule
IdealGPT & 50.7 & 55.3 \\
\midrule
\textbf{SQ-7b} & 32.66 & 48.98  \\
\midrule
\textbf{SQ-13b} & 45.78 & 61.44 \\
\bottomrule
\end{tabular}
}
\vspace{1mm}
\caption{Comparison experiments: Like the paper of IdealGPT did, we sampled 5000 data from each of the VCR and SNLI-VE. \textbf{SQ-7b} is LLaVA-7b-v1.5 fine-tuned by our proposed SQ framework with extra CapQA dataset and \textbf{SQ-13b} is that of LLaVA-13b-v1.5 similarly.}
\vspace{-3mm}
\label{tab:contrast_experiment}
\end{table}

\begin{table}[t!]
\centering
\scalebox{0.68}{
\begin{tabular}{l |lll |ll |ll| llll}
\toprule
\multirow{2}{*}{Method} & QA$_{qg}$ & QA$_{mt}$ & Caption & \multicolumn{2}{c|}{MME} & \multicolumn{2}{c|}{CapQA$^\text{30}$} & \multirow{2}{*}{SQA$^\text{I}$} & \multirow{2}{*}{VQA$^\text{T}$} & \multirow{2}{*}{GQA}  & \multirow{2}{*}{MM-vet} \\
  & & & & Percep & Cog & HalS & QQS & &  & \\
\midrule
\textbf{SQ-caponly}  &&& \checkmark & \underline{1465.4} & \underline{286.4} & 62.9 & 75.8  & \underline{67.63} & \underline{57.67} & \underline{58.51} & 30.0 \\
\textbf{SQ-noqg}   & &\checkmark & \checkmark  & 1421.1 & 282.5 & \underline{68.1} & 75.8  & 67.13 & 55.71 & 55.73 & \underline{30.3} \\
\textbf{SQ}       &\checkmark &\checkmark & \checkmark   & \textbf{1523.4} & \textbf{306.7} & \textbf{69.7} & \textbf{86.7}  & \textbf{68.37} & \textbf{58.57} & \textbf{58.78} & \textbf{31.4} \\
\bottomrule
\end{tabular}
}
\vspace{1mm}
\caption{Ablation experiments on 6 benchmarks. SQA$^\text{I}$: ScienceQA-IMG(zero-shot); VQA$^\text{T}$: TextVQA; CapQA$^{30}$: The first 30 samples of the CapQA evaluation set. SQ-caponly: Retain only the caption portion of the CapQA label during fine-tuning. SQ-noqg: Exclude only the question generation prompt-response pair during fine-tuning. QA$_{qg}$: the first turn QA; QA$_{mt}$: multi-turn QA. The tested VLM is LLaVA-v1.5-7b}
\vspace{-3mm}
\label{tab:contrast_experiment}
\end{table}

\begin{table}[t!]
\centering
\scalebox{0.7}{
\begin{tabular}{l|lll| lll| ll| l}
\toprule
\multirow{2}{*}{Method} & \multicolumn{3}{c|}{POPE(Acc)} & \multicolumn{3}{c|}{POPE(F1)} & \multicolumn{2}{c|}{MME} & \multirow{2}{*}{GQA} \\
 & Rand & Pop & Adv & Rand & Pop & Adv & Percep & Cog & \\
\midrule
LRV-Instruction & 0.86 & 0.73 & 0.65 &  0.65 & 0.79 & 0.73  & 1298.78 & \textbf{328.21} & \textbf{0.64} \\
\midrule
\textbf{SQ}  & \textbf{0.89} & \textbf{0.89} & \textbf{0.85} &  \textbf{0.85} & \textbf{0.86} & \textbf{0.84}  & \textbf{1523.4} & 306.7 & 0.59  \\
\bottomrule
\end{tabular}
}
\vspace{1mm}
\caption{Comparison experiments of LRV-Instruction and SQ. LRV-Instruction: \textit{Liu et al. "Mitigating Hallucination in Large Multi-Modal Models via Robust Instruction Tuning." ICLR 2024}}
\vspace{-3mm}
\label{tab:contrast_experiment}
\end{table}

\begin{table}[t!]
\centering
\scalebox{0.7}{
\begin{tabular}{l|l |lll| lll| lll| l}
\toprule
\multirow{2}{*}{Method} & \multirow{2}{*}{Type} & \multicolumn{3}{c|}{1-th Run} & \multicolumn{3}{c|}{2-th Run} & \multicolumn{3}{c|}{3-th Run} & \multirow{1}{*}{\textbf{Avg Score}} \\
 & & pred/gt & gt  & pred & pred/gt & gt  & pred & pred/gt & gt  & pred & pred/gt \\
\midrule
\textbf{LLaVA-1.5} & HalS & 52.5 & 80.0 & 42.0 & 51.7 & 79.3 & 41.0 & 51.9 & 79.7 & 41.3 & 52.0 \\
\textbf{GPT-4o} & HalS & 72.8 & 76.0 & 55.3 & 74.0 & 77.0 & 57.0 & 74.2 & 76.3 & 56.7 & 73.7 \\
\textbf{SQ} & HalS & 69.7 & 78.0 & 54.3 & 69.4 & 78.3 & 54.3 & 67.9 & 78.0 & 53.0 & 69.0 \\

\midrule
\textbf{LLaVA-1.5} & QQS & 39.2 & 40.0 & 15.7 & 39.2 & 40.0 & 15.7 & 38.6 & 39.7 & 15.3 & 39.0 \\
\textbf{GPT-4o} & QQS  & 97.5 & 40.0 & 39.0 & 97.5 & 40.0 & 39.0 & 96.7 & 40.0 & 38.7 & 97.2 \\
\textbf{SQ} & QQS & 86.7 & 40.0 & 34.7 & 85.8 & 40.0 & 34.3 & 86.7 & 40.0 & 34.7 & 86.4 \\
\bottomrule
\end{tabular}
}
\vspace{1mm}
\caption{Comparison experiments of LLaVA-1.5 and GPT-4o and SQ on the first 30 samples of the CapQA evaluation set. The meanings of the HalS and QQS metrics are the same as defined in the paper. We run the evaluation three times to eliminate randomness and take the average as the final score. For the evaluation, we use GPT-4o-mini-aid.}  
\vspace{-3mm}
\label{tab:contrast_experiment}
\end{table}


\clearpage

\end{document}